\documentclass[a4paper,twoside]{article}

\usepackage{epsfig}
\usepackage{subcaption}
\usepackage{calc}
\usepackage{amssymb}
\usepackage{amstext}
\usepackage{amsmath}
\usepackage{amsthm}
\usepackage{multicol}
\usepackage{pslatex}
\usepackage{apalike}
\usepackage{tabularx}
\newcolumntype{L}{>{\raggedright\arraybackslash}X}
\newcolumntype{C}{>{\centering\arraybackslash}X}
\newcolumntype{R}{>{\raggedleft\arraybackslash}X}
\usepackage{hyperref}

\usepackage{booktabs}

\newcommand{\authornote}[1]{\begin{minipage}[t]{6.2in} \centering \fontsize{9}{11}\selectfont \textit{ #1} \end{minipage} \\}

\newlength{\verticalSpacePreSections}
\newlength{\verticalSpacePostSections}
\newlength{\verticalSpaceParagraphs}
\usepackage{SCITEPRESS}     

\begin{document}

\title{From Explanations to Segmentation: Using Explainable AI for Image Segmentation}

\author{\authorname{Clemens Seibold\sup{*}\sup{1}\orcidAuthor{0000-0002-9318-5934}, Johannes Künzel\sup{*}\sup{1}\orcidAuthor{0000-0002-3561-2758}, Anna Hilsmann\sup{1}\orcidAuthor{0000-0002-2086-0951} and Peter Eisert\sup{1,2}\orcidAuthor{0000-0001-8378-4805}}
\affiliation{\sup{1}Fraunhofer Institute for Telecommunications, Heinrich Hertz Institute, HHI, Einsteinufer 37, 10587 Berlin, Germany}
\affiliation{\sup{2}Visual Computing Group, Humboldt University Berlin, Unter den Linden 6, 10099 Berlin, Germany}
\email{\{clemens.seibold, johannes.kuenzel, anna.hilsmann, peter.eisert\}@hhi.fraunhofer.de}
\authornote{\sup{*}Clemens Seibold and Johannes Künzel have contributed equally.}
}


\abstract{The new era of image segmentation leveraging the power of Deep Neural Nets (DNNs) comes with a price tag: to train a neural network for pixel-wise segmentation, a large amount of training samples has to be manually labeled on pixel-precision.
In this work, we address this by following an indirect solution.
We build upon the advances of the Explainable AI (XAI) community and extract a pixel-wise binary segmentation from the output of the Layer-wise Relevance Propagation (LRP) explaining the decision of a classification network.
We show that we achieve similar results compared to an established U-Net segmentation architecture, while the generation of the training data is significantly simplified.
The proposed method can be trained in a weakly supervised fashion, as the training samples must be only labeled on image-level, at the same time enabling the output of a segmentation mask. This makes it especially applicable to a wider range of real applications where tedious pixel-level labelling is often not possible.}

\onecolumn \maketitle \normalsize \setcounter{footnote}{0} \vfill


\section{\uppercase{Introduction}}
\label{sec:introduction}
\vspace{\verticalSpacePostSections}

Image segmentation describes the demanding task of simultaneously performing object recognition and boundary segmentation and is one of the oldest problems in computer vision \cite[Ch. 5]{Szeliski.2011}.
It is also often a crucial part of many visual understanding systems.

\begin{figure}[!h]
\begin{subfigure}[b]{\linewidth}
     \centering
     \includegraphics[width=\textwidth]{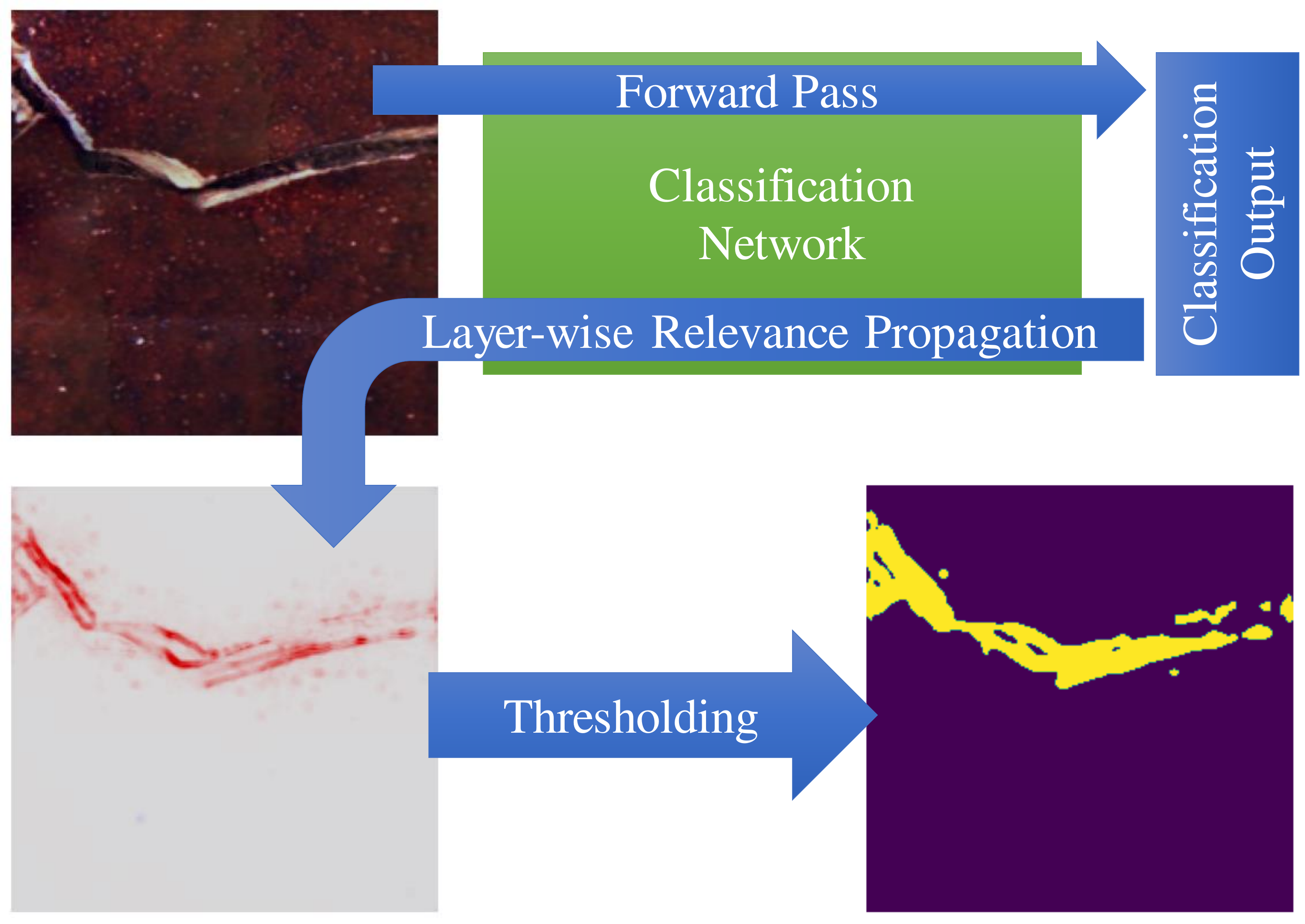}
\end{subfigure}
\caption{Overview of the proposed system. An image containing a crack is passed into a classification network, which learned to separate images with and without cracks. Pixel contributing to the class “with crack” get highlighted using Layer-wise Relevance Propagation (LRP). In consequence, a pixel-wise segmentation is generated without the requirement of pixel-wise labeled training data.}
\label{fig:overview}
\end{figure}

\begin{figure*}[t]
     \centering
     \begin{subfigure}[b]{0.24\textwidth}
         \centering
         \includegraphics[width=\textwidth]{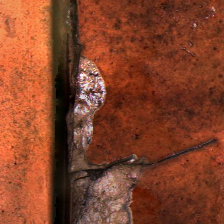}
     \end{subfigure}
     \begin{subfigure}[b]{0.24\textwidth}
         \centering
         \includegraphics[width=\textwidth]{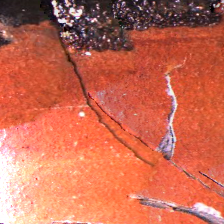}
     \end{subfigure}
     \begin{subfigure}[b]{0.24\textwidth}
         \centering
         \includegraphics[width=\textwidth]{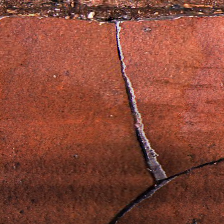}
     \end{subfigure}
     \begin{subfigure}[b]{0.24\textwidth}
         \centering
         \includegraphics[width=\textwidth]{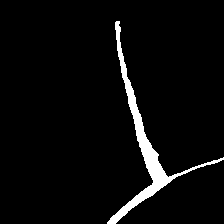}
     \end{subfigure}
        \caption{Examples of images depicting the various appearances of cracks in sewer pipes. The rightmost image shows the segmentation mask of the neighboring image.}
        \label{fig:PipeSamples}
\end{figure*}

\begin{figure*}[ht]
     \centering
     \begin{subfigure}[b]{0.24\textwidth}
         \centering
         \includegraphics[width=\textwidth]{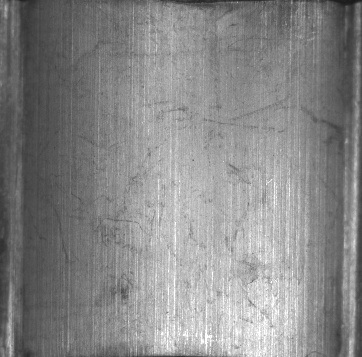}
     \end{subfigure}
     \begin{subfigure}[b]{0.24\textwidth}
         \centering
         \includegraphics[width=\textwidth]{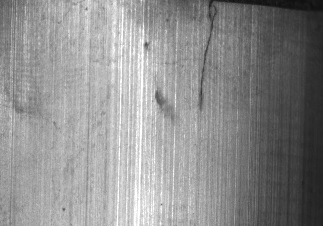}
     \end{subfigure}     
     \begin{subfigure}[b]{0.24\textwidth}
         \centering
         \includegraphics[width=\textwidth]{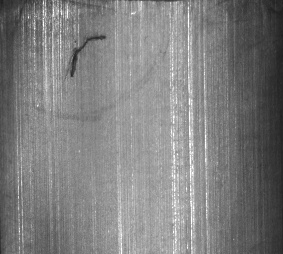}
     \end{subfigure}
     \begin{subfigure}[b]{0.24\textwidth}
         \centering
         \includegraphics[width=\textwidth]{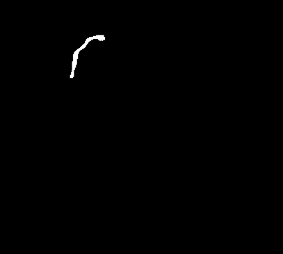}
     \end{subfigure}
        \caption{Examples of images depicting damaged and damage-free magnetic tile surfaces. 
		The left image shows a damage-free magnetic tile and the second image from the left a damaged magnetic tile. The two images on the right show a damaged magnetic tile and its segmentation mask.}
        \label{fig:MagneticSamples}
\end{figure*}

Recent advances of deep learning models resulted in a fundamental change in conjunction with remarkable performance improvements.
However, to train these models, highly accurate labeled data in sufficient large numbers is mandatory.
The goal of our method, depicted in \autoref{fig:overview}, is to circumvent this cumbersome task, by going some extra miles during inference.
To do so, we got inspiration from the field of Explainable AI (XAI) by Layer-wise Relevance Propagation (LRP), presented first by \cite{Bach.2015}.
LRP is usually used to highlight pixels contributing to the decision of a classification network and to get further insights into the decision making process.
In this initial work, we focus on binary semantic segmentation.
We train a VGG-A network architecture to assign an input image to one of two classes.
Afterwards, we use LRP to highlight the pixel contributing to the decision of the network and investigate three segmentation techniques to generate the final segmentation mask.
This idea enables a weakly supervised training of a segmentation method, which needs only image-wise labeled data to train a classification network, but outputs a pixel-wise segmentation mask.
Further, we show that our approach yields comparable results to dedicated segmentation networks, but without the cumbersome requirement of pixel-wise labeled ground-truth data for training.
We put our approach to test on two different datasets - example images can be found in \autoref{fig:PipeSamples} and \autoref{fig:MagneticSamples}.

In the remaining paper, we will summarize related publications in \autoref{sec:relatedwork}, followed by an explanation of our system in \autoref{sec:methods}.
In \autoref{sec:experiments} we evaluate against a classical semantic segmentation network and discuss several design options and their performance impacts.

\vspace{\verticalSpacePreSections}
\section{\uppercase{Related Work}}
\label{sec:relatedwork}
\vspace{\verticalSpacePostSections}

In the past decades, generations of researchers have developed image segmentation algorithms and the literature divides them into three related problems.
\textit{Semantic segmentation} describes the task of assigning a class label to every pixel in the image.
With \textit{instance segmentation}, every pixel gets assigned to a class, along with an object identification.
Thus, it separates different instances of the same class.
\textit{Panoptic segmentation} performs both tasks concurrently, producing a semantic segmentation for “stuff” (e.g.~background, sky) and “things” (objects like house, cars, persons).

There are classical approaches like active contours \cite{Kass.1988}, watershed methods \cite{Vincent.1991} and graph-based segmentation \cite{Felzenszwalb.2004}, to only name a few.
A good overview for these classical approaches can be found in \cite[Ch. 5]{Szeliski.2011}.
With the advance of deep learning techniques in the area of image segmentation, new regions in terms of robustness and accuracy can be reached.
One can find a comprehensive recent review of deep learning techniques for image segmentation in \cite{Minaee.2021}.
This review includes an overview over different architectures, training strategies and loss functions, so we refer the interested reader there, to get a current overview.
All these approaches share one drawback: they need pixel-wise annotated images for training.

In the field of Explainable AI (XAI), several algorithms for the explanation of network decisions have been proposed.
The authors of \cite{Sundararajan2017AxiomaticAF} proposed a method called Integrated Gradients, where gradients are calculated for linear interpolations between the baseline (for example, a black image) and the actual input image.
By averaging over these gradients pixel with the strongest effect on the model’s decision are highlighted.
SmoothGrad \cite{Smilkov2017SmoothGradRN} generates a sensitivity map by averaging over several instances of one input image, each one augmented by added noise.
This way, smoother sensitivity maps can be generated.
There are many more methods and a comprehensive review would be out of scope for this paper, but we refer the interested reader to \cite{Linardatos2021ExplainableAA,BARREDOARRIETA202082,DBLP:series/lncs/11700}.
From this bunch of methods, we selected LRP as it results in comparatively sharp heatmaps and is therefore the best starting point for the generation of segmentation maps. Furthermore, the authors of \cite{Seibold2021} showed that with a simple extension LRP can be utilized to localize traces of forgery in morphed face images.

\vspace{\verticalSpacePreSections}
\section{\uppercase{Methods}}
\label{sec:methods}
\vspace{\verticalSpacePostSections}

\subsection{Overview}
\vspace{\verticalSpacePostSections}

Massive amounts of pixel-wise labeled segmentation masks are usually the training foundation of deep neural networks for image segmentation tasks.
Obtaining them is usually a tedious manual process, introducing intrinsic problems by itself, because of the variance in the annotations and the often fuzzy definitions of object boundaries.
Therefore, we come up with an indirect way.
In our approach, we train a classification network instead of a segmentation network, consequently reducing the annotation work by a great margin and removing the requirement of an exact definition of object boundaries.
To segment an image, we first pass it through the classification network, see \autoref{sec:clas}, which outputs if the object we want to segment is in the image or not.
If the first is true we pass the classification network a second time, but this time from the back to the front, using the LRP technique described in \autoref{subsec:lrp} \cite{Bach.2015}, which is an XAI technique, highlighting the pixel contributing to the DNN's decision in a heatmap. We use this heatmap in order to generate a segmentation mask without the cumbersome task of manual pixel-wise labeling.

\subsection{Network Architectures}
\vspace{\verticalSpacePostSections}

\subsubsection{Classification}
\label{sec:clas}
\vspace{\verticalSpacePostSections}
For binary classification, we resort to the classical VGG-11 architecture without batch normalization, as described in \cite{Simonyan2015VeryDC}, but with only 128 neurons in each of the fully connected layers.
We use this architecture, as it is readily available, well understood and a good starting point for the use of the LRP framework \cite{Bach.2015}.
For further improvements of the segmentation results generated with LRP, we also adapt the VGG-11 architecture, connecting the outputs of the last max-pooling layer directly to the two output neurons.
This facilitates the classification accuracy, as well as the indirect segmentation (see \autoref{sec:experiments}).
For both configurations, we start the training with pretrained weights for the convolutional and randomly initialized weights for the fully connected layers.
We use a learning rate of 0.001 and 0.0001 for the fully connected layers and the refined convolutional layers, respectively.

\subsubsection{Native Segmentation}
\vspace{\verticalSpacePostSections}

For a comparison of our proposed method against an established network architecture for image segmentation, we have chosen the U-Net architecture as described in \cite{Ronneberger.2015}, as it was developed especially for very small datasets (the authors of the original work used only 30 training samples).
We train the network, as described in the original work, in a classical supervised way using data augmentation techniques like affine transformations and random elastic deformations to cope with the small training datasets of 60 and 119 for the sewer pipes and the magnetic tiles, respectively.

\subsection{LRP}
\label{subsec:lrp}
\vspace{\verticalSpacePostSections}

\subsubsection{Principles}
\vspace{\verticalSpacePostSections}

Layer-wise Relevance Propagation (LRP) \cite{Bach.2015} is an interpretability method for DNNs.
It was designed to highlight on a pixel-level the structures in an image that are relevant for the DNN’s decision. 
To this end, it assigns to each input neuron of a DNN, e.g.~each pixel of an image, a relevance score that reflects its impact on the activation of a class of interest.
A positive relevance score denotes a contribution to the activation, while a negative relevance score denotes an inhibition.

In a first step, LRP assigns a relevance value to a starting neuron that represents the class of interest.
Given this initialization, LRP propagates this starting relevance backwards through the DNN into the input image.
To this end, it iterates from the last layer to the input image through all layers. 
In each iteration step, it assigns relevance scores to neurons in the current layer based on their activations and weights to neurons in the subsequent layer and the neurons’ relevance scores in the subsequent layer. 
If a neuron contributes to an activation of a neuron in the subsequent layer, it receives a percentage of its relevance. 
If it inhibits the activation, it gets a negative relevance percentage.
LRP can make use of different rules for the propagation of relevance from neurons in one layer to the neurons in its predecessor.
The rules define how LRP propagates the relevance in every single step and consider various properties of the activations and connections in different parts of the DNN. 
\cite{Montavon2019} shows that the most accurate and understandable relevance distributions can be achieved by used different rules for different parts of a DNN.


\begin{table}[ht]
\caption{Overview of the two different LRP rule sets, which are compared. Each row contains a specific LRP rule and the layers in the VGG-A architecture where they are used.}
\label{tab:LRP_Rules}
\centering
\begin{tabularx}{\linewidth}{@{}rCC@{}}
\toprule
\textbf{Rule names}       & \textbf{LRP Ours}                                                                            & \textbf{LRP Montavon}                                                             \\ \midrule
$z^B$                       & conv1\_1                                                                                     & conv1\_1                                                                          \\ \midrule
$\alpha$-$\beta$            & \begin{tabular}[c]{@{}c@{}}conv2\_1\\ conv3\_1\end{tabular}                                  &                                                                                   \\ \midrule
$\gamma$                    & \begin{tabular}[c]{@{}c@{}}conv3\_2\\ conv4\_1\\ conv4\_2\\ conv5\_1\\ conv5\_2\end{tabular} & \begin{tabular}[c]{@{}c@{}}conv2\_1\\ conv3\_1\\ conv3\_2\end{tabular}            \\ \midrule
$\epsilon$                  & FC                                                                                           & \begin{tabular}[c]{@{}c@{}}conv4\_1\\ conv4\_2\\ conv5\_1\\ conv5\_2\end{tabular} \\ \midrule
$0$                         &                                                                                              & FC                                                                                \\ \bottomrule
\end{tabularx}
\end{table}

In our experiments, we use two different sets of LRP rules. An overview is given in \autoref{tab:LRP_Rules}.
The parameters for the rules are $\epsilon=0.25std$ and $\gamma=0.25$.
In both cases, we use the LRP-$z^B$ rule for the first layer, which maps the relevance into the image.
The LRP-$z^B$ rule considers that it has to propagate the relevance to pixels that contain real values and not ReLU activations like the neurons in the DNN.
The first combination of rules and parameters has been shown to be suitable for VGG-like architectures \cite{Montavon2019}.
While the LRP-0 rule is close to the activation function of the network, the LRP-$\epsilon$ rule focuses on more salient features and the LRP-$\gamma$ rule is most suitable to spread the relevance to the whole feature.
Empirically, we found a suitable second combination of rules, which we included in our experiments.
It uses the LRP-$\epsilon$ rule for the fully connected layers to focus already here on more salient features. The use of the LRP-$\gamma$ rule for the middle layers enforces an early spread of the relevance to the whole feature. The LRP-$\alpha$-$\beta$ rule with $\alpha=2$ in the lower layers considers inhibiting and contributing features differently and puts a strong focus on contributing activations. This rule leads to more balanced relevance maps with strong focus on inhibiting regions. 
For further details on LRP and the characteristics of its different relevance propagation rules, we refer to \cite{kohlbrenner-ijcnn20}.

\subsubsection{Segmentation from LRP}
\label{subsubsec:segmentation}
\vspace{\verticalSpacePostSections}

LRP assigns to each pixel a relevance score with values within an arbitrary interval. 
In order to transfer these relevance distributions to a segmentation map, we developed three different approaches. 
The first one \textit{Simple} is based on the simple automatic thresholding algorithm described in \cite{Umbaugh.2017}.
The other two approaches, \textit{GMM} and \textit{BMM}, are based on Mixture Models and optimized using an 
Expectation-Maximization algorithm.

\vspace{\verticalSpaceParagraphs}
\paragraph{Simple}
To calculate the segmentation of foreground and background, the LRP activations are normalized first and the mean defines the initial threshold over all the activations.
Based on the initial threshold, the image can be separated into foreground and background and the mean over all values in both classes is calculated.
The new threshold then arises from the average of both mean values.
This iterative refinement stops if the threshold value converges.

\vspace{\verticalSpaceParagraphs}
\paragraph{GMM}
This segmentation method is based on a Gaussian Mixture Model (GMM) to distinguish between relevant regions (damages) and background. 
In a first step, we apply a 2D mean filter with a dimension of five by five on the relevance map. 
This smooths out extreme relevance peaks for single pixels and makes the relevance distribution in the inner part of a damage smoother.
Our GMM has three components and is fit to the relevance distribution considering only the 1-D relevance scores and no spatial information. 
We used python’s scikit-learn package \cite{pedregosa2011scikit} to fit the GMM to the data. We initialize the GMM using the k-Means algorithm to find first belongings of the samples to the distributions and thus to initialize the parameters. 
In order to identify the component of the GMM that describes the relevance values covering the damages, we selected the component with the largest likelihood for the maximal relevance value. 
The final segmentation map is calculated by selecting all pixels that belong to this component with a likelihood of more than 50\%. 

\vspace{\verticalSpaceParagraphs}
\paragraph{BMM}
Our Beta Mixture Model for segmentation consists of two Beta-distributions. See \eqref{eq:betaDist} for the definition of the probability density function of the Beta-distribution. 
\begin{equation}\label{eq:betaDist}
    f(x; \alpha, \beta) = \frac{1}{B(\alpha, \beta)}x^{\alpha-1}(1-x)^{\beta-1},
\end{equation}
with $B(\cdot)$ being the normalization factor as defined in \eqref{eq:betaDistNorm}
\begin{equation}\label{eq:betaDistNorm}
	B(\alpha, \beta) = \dfrac{\Gamma(\alpha)\Gamma(\beta)}{\Gamma(\alpha+\beta)}
\end{equation}
and $\Gamma(\cdot)$ is the Gamma function. The idea of this model is to use two skewed distributions to describe the relevance scores. One distribution characterizes the large amount of background pixels with low positive relevance values and the other one the areas containing damages with large relevance values. The distributions are fit using an EM-algorithm with outlier removal and weights for the relevance distributions. The details are described in the following.

First, the relevance maps are filtered as in our GMM segmentation approach. In a next step, we set all negative relevance values to zero and remove 50\% of the smallest relevance values, since the damages cover in all cases significantly less pixels than 50\% of the image. 
Subsequently, the relevance scores are normalized to the interval [0,1], since the Beta-distribution is defined only on this interval. To this end, we map the smallest value to zero and the largest to one using an affine transformation. 
The BMM is fit to these processed data using an EM-algorithm with the following modifications in the expectation step.

Pixels with a larger relevance value than the probability density function’s mean of the component that represents the damages are assigned with a probability of 100\% to this component. 
Pixels that are within the 90\% of the lower sided confidence interval of the component that represents the background are assigned with a probability of 100\% to this component. Finally, we weight the probabilities of each component by the sum of the component’s probabilities. 
These modifications in the maximization step avoid that the large number of small relevance values from the background pixels affects the component that describes the relevance values of the damaged areas and compensates the big differences in the number of relevant (damage) and non-relevant (background) pixels.

\subsection{Datasets}
\vspace{\verticalSpacePostSections}

To showcase the applicability of our proposed method, we work with two different datasets in our experiments, which we describe in detail below.

\vspace{\verticalSpaceParagraphs}
\paragraph{Cracks in Sewer Pipes}
Sewer pipe assessment is usually done with the help of mobile robots equipped with cameras.
In our case, the robot was equipped with a fisheye lens, resulting in severe image distortions.
Therefore, we performed a preprocessing of the original footage, as described in \cite{Kuenzel2018}, consisting of camera tracking, image reprojection and enhancement.
For the classification of damaged and undamaged pipe surfaces, we manually cropped 628 and 754 images respectively, with a size of 224 by 224 pixels.
The damages show a huge variety in size, color and shape, as can be seen in \autoref{fig:PipeSamples}.
During the training of the classification network, we performed no further data augmentation.
For the training of a classical segmentation network, we cropped another set of images containing pipe cracks and manually created the segmentation masks.
We divided the dataset into a testing and validation dataset, each with 20\% of all images, and a training set with the remaining 60\% of all images.
During the training of the segmentation network, we used elastic deformations and affine transformations for data augmentation \cite{Ronneberger.2015}.
The testing and validation data are augmented by adding the horizontally and vertically flipped version of each image to the corresponding set.

\vspace{\verticalSpaceParagraphs}
\paragraph{Cracks in Magnetic Tiles}
As a second dataset, we use the magnetic tile defect datasets provided by the authors of \cite{Huang2018}.
This set contains 894 images of magnetic tiles without any damage and 190 images of magnetic tiles with either a crack or a blowhole. 
These damages are very small and cover only a few percent of an image. 
The images in this dataset are of different sizes with widths between 103 and 618 pixels and heights between 222 and 403 pixels. 
We manually cropped all images with damaged magnetic tiles, such that each random crop of a 224 by 224 pixels large region contains the damage. 
During training, the images are randomly cropped to a size of 224 by 224 pixels, while during testing and validation we cropped the center of the images. 
Images with a height or width smaller than 224 have been rescaled to reach the minimal size of 224 pixels. 
We divided the dataset into a testing and validation dataset, each with 20\% of all images, and a training set with the remaining 60\% of all images.
We ensured that this distribution holds also for all damage types and damage free images. 
When splitting the images into these sets, we considered that the authors of the dataset captured each damage and damage-free region up to six different times and split the images such that each damage or damage-free region area is in one set only.
We augmented the data during training using random cropping and random horizontal and vertical flipping.
The testing and validation data are augmented by adding the horizontally and vertically flipped version of each image to the corresponding set.
Example images can be found in \autoref{fig:MagneticSamples}.

\vspace{\verticalSpacePreSections}
\section{\uppercase{Experiments}}
\label{sec:experiments}
\vspace{\verticalSpacePostSections}

We tested the proposed LRP-based segmentation methods on the Sewer Pipe Cracks and the Magnetic Tile Cracks dataset and compared their performance with the segmentation results from a U-Net. We evaluated all combinations of our three proposed threshold estimation methods, LRP rules and VGG-based networks to study their suitability for image segmentation.
The used evaluation metrics are Intersection over Union (IoU) and the two for binary classification tasks common metrics, precision and recall.
Since this is a binary problem, we calculated all metrics only for the pixels segmented as damage.
Furthermore, we analyzed the Precision-Recall (PR) Curves of the different segmentation approaches.
Whereas the U-Net, the BMM approach and the GMM approach output a value for the likelihood that a pixel shows a damaged area, the simple approach outputs only a binary decision and no PR curve can be calculated for this approach.
In exchange for the simple approach, we calculated the PR curve on the LRP-output after mapping it into the interval [0,1] using an affine transformation.

The PR curves contain striking horizontal lines, which origin is explained in the following.
The GMM approach assigns to a large number of pixels a likelihood of one for being a damage. 
Two components of the GMM have a mean around zero and very small variances. 
The third component, which describes the damage, has a significantly larger variance and mean. 
Due to the narrow peak of the first two components that describe non-damage pixels, their probability density functions are already zero for moderate relevance scores, when using 64 bit floating point numbers as defined in IEEE 754-2008. 
Especially, for the Magnetic Tile dataset, the contrast of the large amount of background pixels with relevance scores close to zero and the small amount of pixels showing damaged areas with large relevance scores causes this behavior. 
It can also be observed for the Sewer Pipes dataset, but to a smaller extent.
Increasing the threshold can thus not improve the precision or change the recall. 
We depicted this point of saturation with a horizontal line.

\subsection{Magnetic Tiles}
\vspace{\verticalSpacePostSections}
Both VGG-A-based networks achieve in all cases a good performance in detecting damages and yield a balanced accuracy of more than 95\%, see Table \ref{tab:AccuMagnetic}.

\begin{table}[h]
\caption{Magnetic Tile Damage Detection Metrics}
\label{tab:AccuMagnetic}
\centering
\begin{tabularx}{\linewidth}{@{}cCcc@{}}
\toprule
             & Balanced Accuracy & TPR    & TNR    \\ \cmidrule(l){2-4} 
VGG-A 128    & 96.1\%            & 95.7\% & 96.5\% \\
VGG-A one FC & 98.5\%            & 97.1\% & 99.9\% \\ \bottomrule
\end{tabularx}
\end{table}

\begin{figure}[h]
    \centering
    \includegraphics[width=0.9\linewidth]{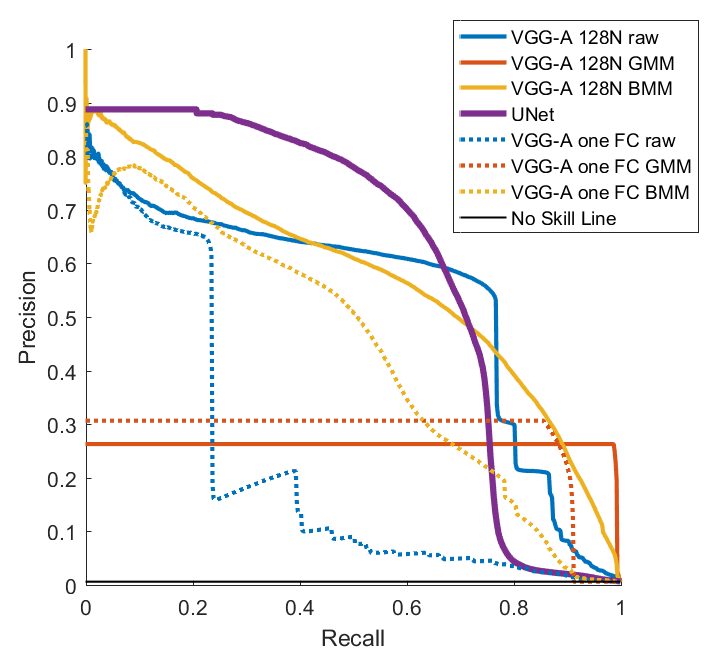}
    \caption{Precision Recall Curves for Damage Segmentation in Magnetic Tile Images with the LRP ruleset from Montavon.}
    \label{fig:PRMagneticTiles_Montavon}
\end{figure}
\begin{figure}[h]
    \centering
    \includegraphics[width=0.9\linewidth]{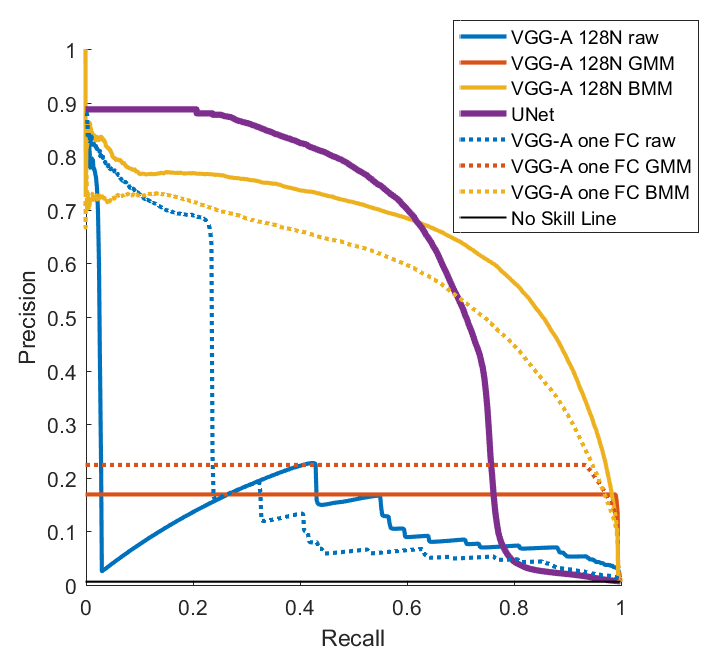}
    \caption{Precision Recall Curves for Damage Segmentation in Magnetic Tile Images with our LRP ruleset.}
    \label{fig:PRMagneticTiles_Ours}
\end{figure}

Table \ref{tab:MetricsMagneticTiles} shows that some of our LRP-based segmentation approaches perform as well as the U-Net segmentation, which requires a pixel-wise segmentation for training.
The performance of our proposed approaches differs strongly in terms of IoU, precision and recall. 
The GMM approach has the worst IoU and precision but by far the best recall. 
The recall of the simple approach is in general better than for the BMM approach, but the BMM provides a better precision. 
There is no model that outperforms all others in all metrics. 
Which approach to choose depends on whether the detector should focus on sensitivity or specificity.

The precision-recall curves in \autoref{fig:PRMagneticTiles_Montavon} and \autoref{fig:PRMagneticTiles_Ours} show that the different results for the metrics in Table \ref{tab:MetricsMagneticTiles} for U-Net, BMM and GMM are not only a matter of threshold, but the approaches perform differently depending on the selected theshold. 
All our approaches can achieve better results than a no-skill segmentation system with a precision of 0.006. 
The BMM approach performs in nearly all, except for a very high recall rates, better than the GMM approach. 
It outperforms the U-Net based segmentation in the recall interval of roughly 0.7 to 1 in the best setting with the \textit{VGG-A 128N DNN} and our proposed LRP-ruleset. 
In the remaining range of 0 to 0.7 its precision is on average only 0.1 worse than the U-Net segmentation. 
In all cases, the GMM model reaches very fast a point of saturation with final precision scores between about 0.15 and 0.3, depending on the used LRP-ruleset and DNN model. 
An explanation for this saturation can be found in the beginning of \autoref{sec:experiments}.

Whether our proposed LRP-ruleset or the ruleset proposed in \cite{Montavon2019} is more suitable for a LRP-based segmentation depends on the approach used for the final segmentation step.
While a raw relevance intensity based and GMM-based segmentation approach yields in general better results with the ruleset  proposed in \cite{Montavon2019}, the best results are achieved using our proposed ruleset and the BMM approach using the \textit{VGG-A 128N} model.

\begin{figure*}[ht]
    \centering
    \newlength{\widthcolsubfigure}
    \setlength{\widthcolsubfigure}{0.14\textwidth}
    \newlength{\hspacevalue}
    \setlength{\hspacevalue}{0.5cm}
    \newlength{\vspacevalue}
    \setlength{\vspacevalue}{0.1cm}
    \begin{subfigure}[b]{\widthcolsubfigure}
        \centering
        \includegraphics[width=\textwidth]{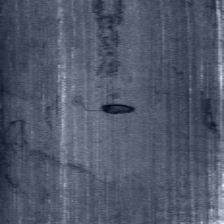}\\
        \vspace{\vspacevalue}
        \includegraphics[width=\textwidth]{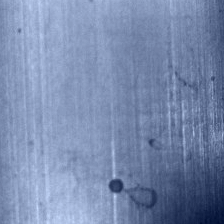}\\        
        \vspace{\vspacevalue}
        \includegraphics[width=\textwidth]{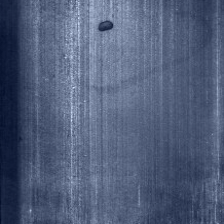}\\
        \vspace{\vspacevalue}
        \includegraphics[width=\textwidth]{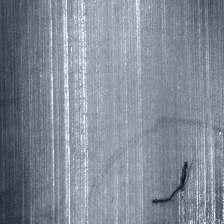}
        \caption{Input images}
    \end{subfigure}    
    \hspace{\hspacevalue}
    \begin{subfigure}[b]{\widthcolsubfigure}
        \centering
        \includegraphics[width=\textwidth]{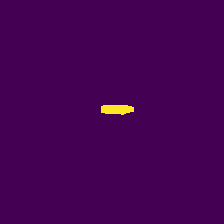}\\
        \vspace{\vspacevalue}
        \includegraphics[width=\textwidth]{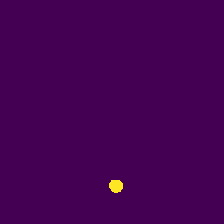}\\
        \vspace{\vspacevalue}
        \includegraphics[width=\textwidth]{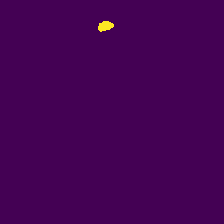}\\
        \vspace{\vspacevalue}
        \includegraphics[width=\textwidth]{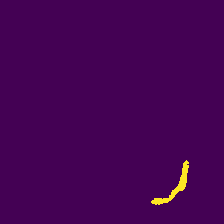}
        \caption{Groundtruth}
    \end{subfigure}      
    \hspace{\hspacevalue}
    \begin{subfigure}[b]{\widthcolsubfigure}
        \centering
        \includegraphics[width=\textwidth]{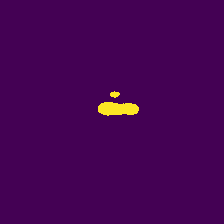}\\
        \vspace{\vspacevalue}
        \includegraphics[width=\textwidth]{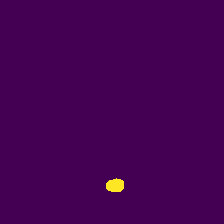}\\
        \vspace{\vspacevalue}
        \includegraphics[width=\textwidth]{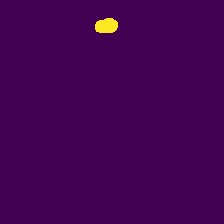}\\
        \vspace{\vspacevalue}
        \includegraphics[width=\textwidth]{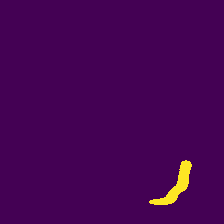}
        \caption{U-Net Seg.\label{fig:magnetic_tiles_segmentation_unet}}
    \end{subfigure}
    \hspace{\hspacevalue}
    \begin{subfigure}[b]{\widthcolsubfigure}
        \centering
        \includegraphics[width=\textwidth]{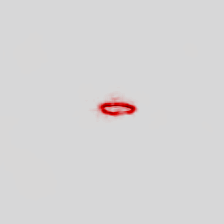}\\
        \vspace{\vspacevalue}
        \includegraphics[width=\textwidth]{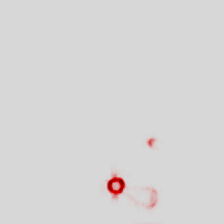}\\
        \vspace{\vspacevalue}
        \includegraphics[width=\textwidth]{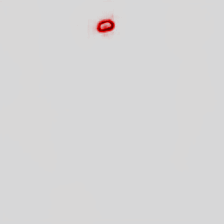}\\
        \vspace{\vspacevalue}
        \includegraphics[width=\textwidth]{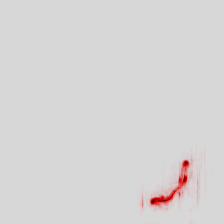}
        \caption{LRP heatmap\label{fig:magnetic_tiles_seg_vgg_128_ours_heatmap}}
    \end{subfigure}
    \hspace{\hspacevalue}
    \begin{subfigure}[b]{\widthcolsubfigure}
        \centering
        \includegraphics[width=\textwidth]{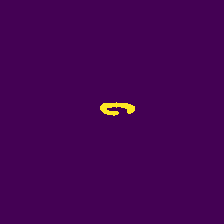}\\
        \vspace{\vspacevalue}
        \includegraphics[width=\textwidth]{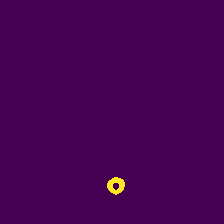}\\
        \vspace{\vspacevalue}
        \includegraphics[width=\textwidth]{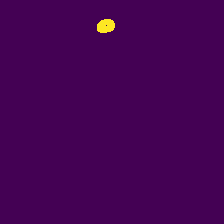}\\
        \vspace{\vspacevalue}
        \includegraphics[width=\textwidth]{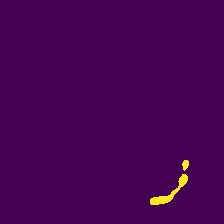}
        \caption{LRP Seg.\label{fig:magnetic_tiles_seg_vgg_128_ours_bmm}}
    \end{subfigure}
    \caption{Example results for segmentations of damages in magnetic tiles generated with U-Net and our proposed LRP-ruleset and the VGG-A 128N architecture.}
    \label{fig:MagneticResults}
\end{figure*}

Figure \ref{fig:MagneticResults} depicts examples for the damage segmentation of magnetic tile images using U-Net as well as our proposed approach. 
The LRP-results in the first two rows show a typical weakness of LRP-based segmentations. 
The relevance scores at the borders of small damages are very large, but inside the defect, they are small and close to zero. 
Thus, a non-sensitive approach does not segment the inner part of a damaged area as such. 
The LRP-based segmentation in the first row does not contain the complete contour of the damage, which is caused by smaller relevance values at one part of the damage’s border. 
This problem can be solved by adjusting the threshold for the segmentation to make the approach more sensitive. 
The example in the bottom row shows that our LRP-based approach is also suitable to segment more complex structures than the ellipsoids showed in the other three examples.
In general, the LRP-based and U-Net Segmentation results can describe the position as well as the shape of the damage with a visually comprehensible precision. 

\begin{table*}[t]
\caption{Overview of the metrics for the magnetic tile damage segmentation. On the left is our proposed LRP rule set and on the right is the one proposed by \cite{Montavon2019}. On each side, the default VGG-A architecture and the architecture with only one fully connected layer are compared. For easier comparison, the U-Net results are listed on both sides.}
\label{tab:MetricsMagneticTiles}
\centering
\begin{tabularx}{\textwidth}{ccccLcccc}
\cmidrule[\heavyrulewidth](r){1-4} \cmidrule[\heavyrulewidth](l){6-9}
\multicolumn{4}{c}{\textbf{LRP Ours}}                                                                                      &  & \multicolumn{4}{c}{\textbf{LRP Montavon}}                                                                                  \\ \cmidrule[\lightrulewidth](r){1-4} \cmidrule[\lightrulewidth](l){6-9} 
\textbf{VGG-A 128N}                 & \multicolumn{1}{l}{IoU} & \multicolumn{1}{l}{Precision} & \multicolumn{1}{l}{Recall} &  & \textbf{VGG-A 128N}                 & \multicolumn{1}{l}{IoU} & \multicolumn{1}{l}{Precision} & \multicolumn{1}{l}{Recall} \\ \cmidrule(lr){2-4} \cmidrule(l){7-9} 
\multicolumn{1}{c|}{Simple}         & 0.349                   & 0.352                         & 0.976                      &  & \multicolumn{1}{c|}{Simple}         & 0.466                   & 0.497                         & 0.897                      \\
\multicolumn{1}{c|}{GMM}            & 0.129                   & 0.129                         & 0.993                      &  & \multicolumn{1}{c|}{GMM}            & 0.216                   & 0.216                         & 0.996                      \\
\multicolumn{1}{c|}{BMM}            & 0.462                   & 0.638                         & 0.679                      &  & \multicolumn{1}{c|}{BMM}            & 0.332                   & 0.703                         & 0.448                      \\ \cmidrule[\lightrulewidth](r){1-4} \cmidrule[\lightrulewidth](l){6-9} 
\textbf{VGG-A one FC}               & IoU                     & \multicolumn{1}{l}{Precision} & \multicolumn{1}{l}{Recall} &  & \textbf{VGG-A one FC}               & IoU                     & \multicolumn{1}{l}{Precision} & \multicolumn{1}{l}{Recall} \\ \cmidrule(lr){2-4} \cmidrule(l){7-9} 
\multicolumn{1}{c|}{Simple}         & 0.386                   & 0.404                         & 0.944                      &  & \multicolumn{1}{c|}{Simple}         & 0.425                   & 0.491                         & 0.808                      \\
\multicolumn{1}{c|}{GMM}            & 0.168                   & 0.169                         & 0.991                      &  & \multicolumn{1}{c|}{GMM}            & 0.238                   & 0.246                         & 0.947                      \\
\multicolumn{1}{c|}{BMM}            & 0.402                   & 0.669                         & 0.614                      &  & \multicolumn{1}{c|}{BMM}            & 0.317                   & 0.686                         & 0.470                      \\ \cmidrule[\heavyrulewidth](r){1-4} \cmidrule[\heavyrulewidth](l){6-9} 
\multicolumn{1}{c|}{\textbf{U-Net}} & 0.461                   & 0.578                         & 0.699                      &  & \multicolumn{1}{c|}{\textbf{U-Net}} & 0.461                   & 0.578                         & 0.699                      \\ \cmidrule[\heavyrulewidth](r){1-4} \cmidrule[\heavyrulewidth](l){6-9} 
\end{tabularx}
\end{table*}

\subsection{Sewer Pipes}
\vspace{\verticalSpacePostSections}

\begin{figure}[h]
    \centering 
    \includegraphics[width=0.9\linewidth]{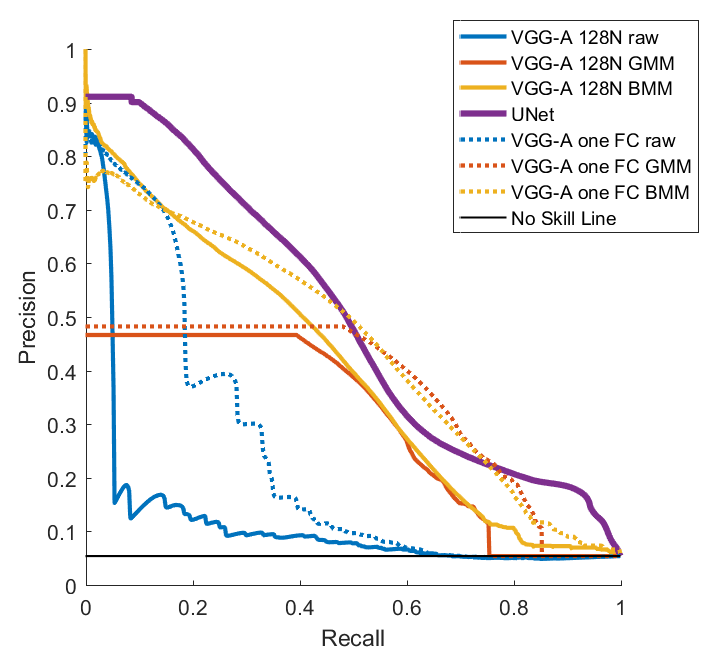}
    \caption{Precision Recall Curves for Damage Segmentation in Sewer Pipe Images with the LRP rule set from Montavon.\label{fig:PRSewerPipes_Montavon}}
\end{figure}
\begin{figure}[h]
    \centering
    \includegraphics[width=0.9\linewidth]{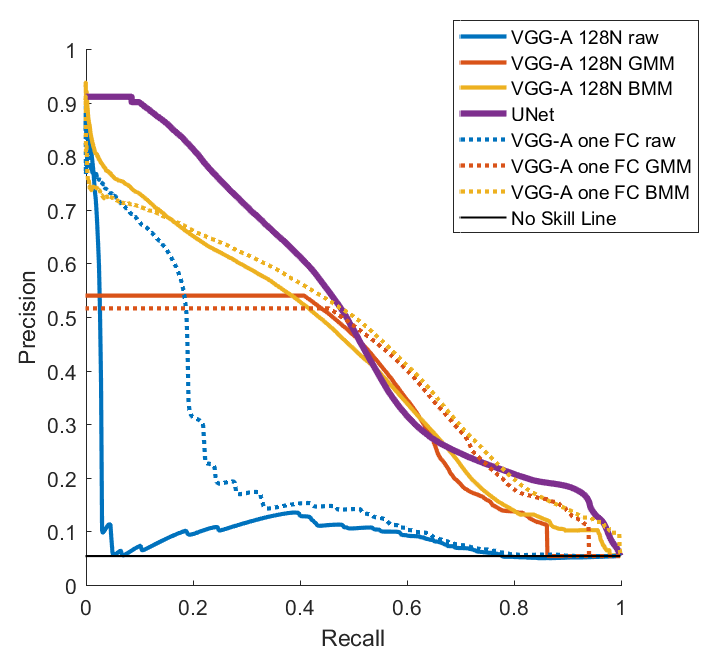}
    \caption{Precision Recall Curves for Damage Segmentation in Sewer Pipe Images with our LRP rule set.\label{fig:PRSewerPipes_Ours}}
\end{figure}

The VGG-A-based classification networks yield more than 95\% balanced accuracy for both network configurations.
By directly connecting the convolutional layers to only two output neurons, a slight performance increase could be achieved, as can be seen in \autoref{tab:AccuSewer}.

\begin{table}[h]
\caption{Sewer Pipe Cracks Detection Metrics}
\label{tab:AccuSewer}
\centering
\begin{tabularx}{\linewidth}{@{}cCcc@{}}
\toprule
             & Balanced Accuracy & TPR    & TNR    \\ \cmidrule(l){2-4} 
VGG-A 128    & 95.3\%            & 92.8\% & 97.5\% \\
VGG-A one FC & 97.4\%            & 98.5\% & 96.2\% \\ \bottomrule
\end{tabularx}
\end{table}

\autoref{tab:MetricsSewerPipes} contains the segmentation metrics for the Sewer Pipe Cracks dataset.
From our approaches, the BMM-based segmentation has the highest IoU values and also outperforms U-Net, when used in conjunction with \textit{VGG-A one FC}.
This configuration also has the highest precision, but lower recall values.
The usage of the GMM-based approach shows similar results compared to simple thresholding algorithm.

The precision-recall curves for our LRP configuration and the one from \cite{Montavon2019} are plotted in \autoref{fig:PRSewerPipes_Ours} and \autoref{fig:PRSewerPipes_Montavon}.
As can be seen from the figures, the utilization of the raw LRP output is not practicable, since the segmentation is barely better than a no-skill segmentation system with a precision of 0.05379, which is just the proportion of pixel labeled as cracks (represented by the black horizontal line).

The utilization of the proposed GMM and BMM based segmentations improves the results by a great margin.
However, the GMM based approaches exhibit a straight horizontal line, for which an explanation can be found in the beginning of \autoref{sec:experiments}.
In the interval between a recall of roughly 0.5 and 0.7 our approach outperforms U-Net, but falls behind for lower recall values.
For recall values greater than 0.7, \textit{VGG-A one FC} and \textit{U-Net} show a similar performance when utilizing our LRP ruleset.
The usage of \textit{LRP Montavon} results in a wider margin and also causes a worse performance for the \textit{VGG-A 128N} configurations, which therefore never exceed the performance of \textit{U-Net}.

\begin{figure*}[ht]
    \centering
    \newlength{\jwidthcolsubfigure}
    \setlength{\jwidthcolsubfigure}{0.14\textwidth}
    \newlength{\jhspacevalue}
    \setlength{\jhspacevalue}{0.5cm}
    \begin{subfigure}[b]{\jwidthcolsubfigure}
        \centering
        \includegraphics[width=\textwidth]{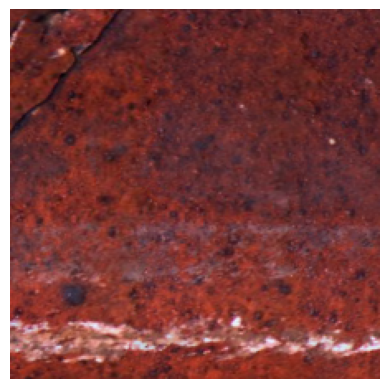}\\        
        \includegraphics[width=\textwidth]{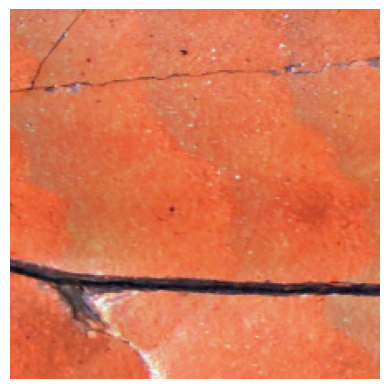}\\        
        \includegraphics[width=\textwidth]{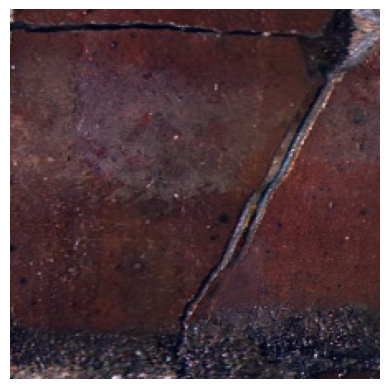}\\
        \includegraphics[width=\textwidth]{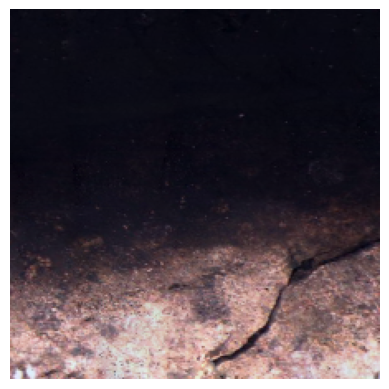}
        \caption{Input images}
    \end{subfigure}    
    \hspace{\jhspacevalue}
    \begin{subfigure}[b]{\jwidthcolsubfigure}
        \centering
        \includegraphics[width=\textwidth]{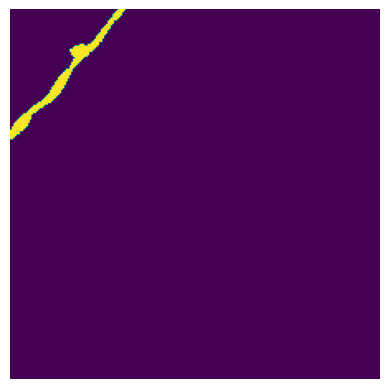}\\
        \includegraphics[width=\textwidth]{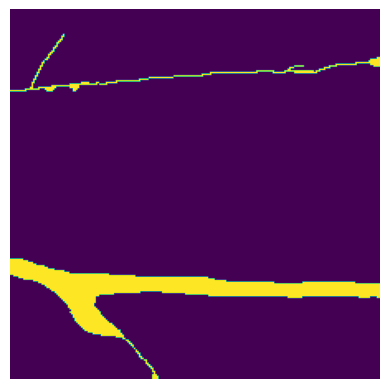}\\
        \includegraphics[width=\textwidth]{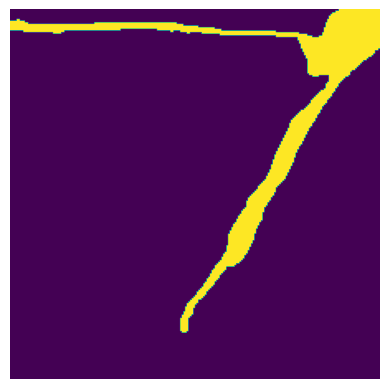}\\
        \includegraphics[width=\textwidth]{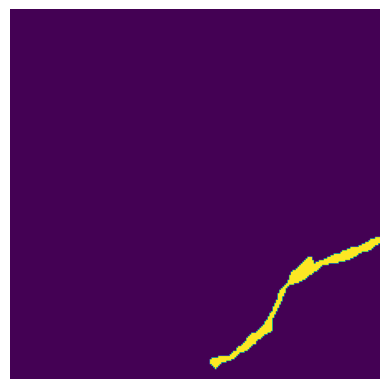}
        \caption{Groundtruth}
    \end{subfigure}      
    \hspace{\jhspacevalue}
    \begin{subfigure}[b]{\jwidthcolsubfigure}
        \centering
        \includegraphics[width=\textwidth]{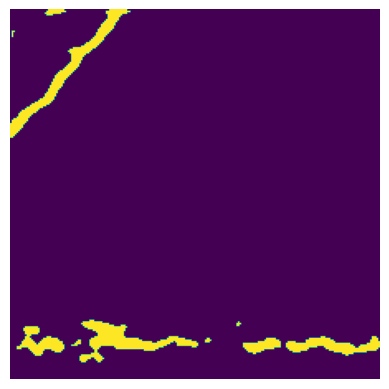}\\
        \includegraphics[width=\textwidth]{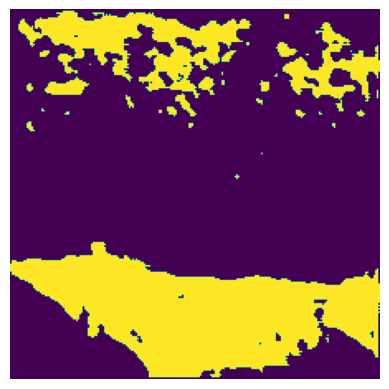}\\
        \includegraphics[width=\textwidth]{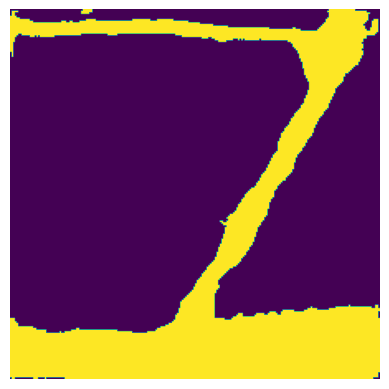}\\
        \includegraphics[width=\textwidth]{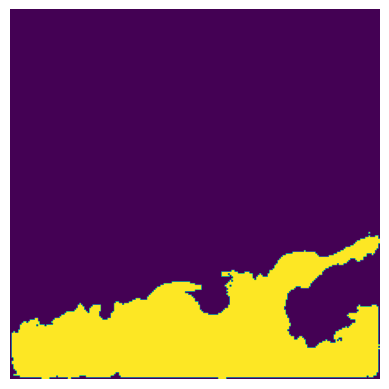}
        \caption{U-Net Seg.\label{fig:segmentation_unet}}
    \end{subfigure}
    \hspace{\jhspacevalue}
    \begin{subfigure}[b]{\jwidthcolsubfigure}
        \centering
        \includegraphics[width=\textwidth]{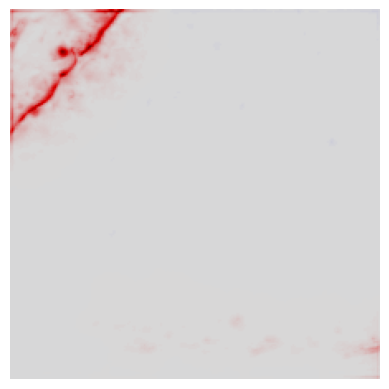}\\
        \includegraphics[width=\textwidth]{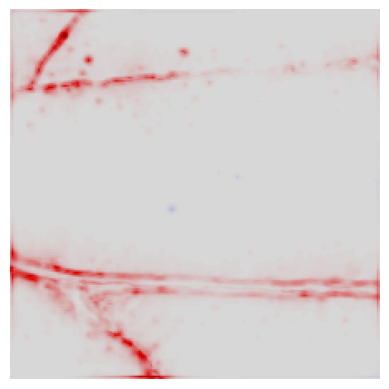}\\
        \includegraphics[width=\textwidth]{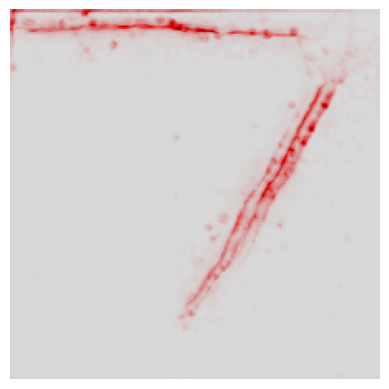}\\
        \includegraphics[width=\textwidth]{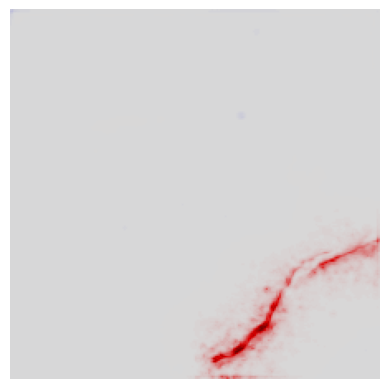}
        \caption{LRP heatmap\label{fig:seg_vgg_one_ours_heatmap}}
    \end{subfigure}
    \hspace{\jhspacevalue}
    \begin{subfigure}[b]{\jwidthcolsubfigure}
        \centering
        \includegraphics[width=\textwidth]{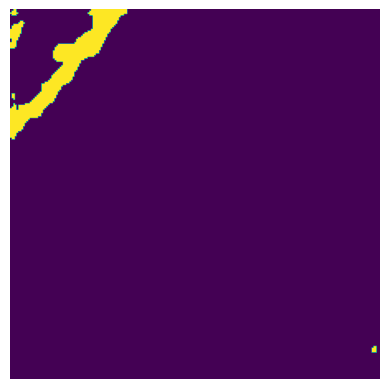}\\
        \includegraphics[width=\textwidth]{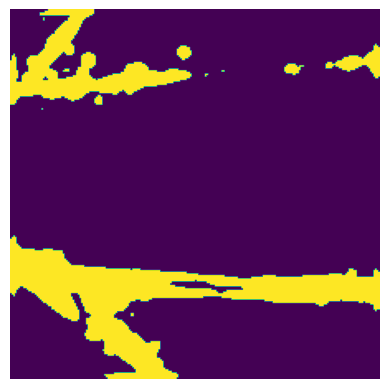}\\
        \includegraphics[width=\textwidth]{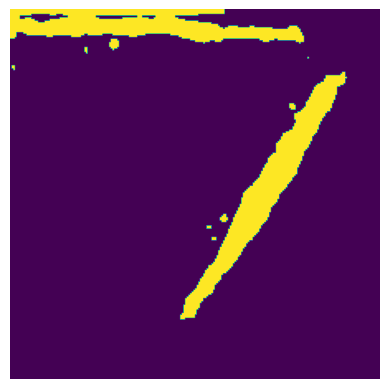}\\
        \includegraphics[width=\textwidth]{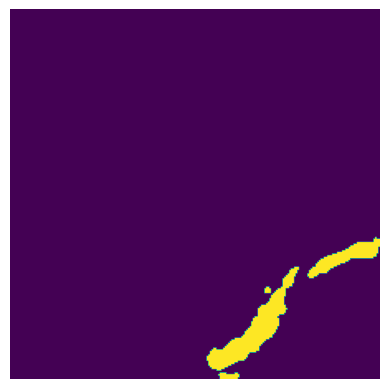}
        \caption{LRP Seg.\label{fig:seg_vgg_one_ours_bmm}}
    \end{subfigure}
    \caption{Example results for segmentations of cracks in sewer pipes generated with U-Net and our best performing configuration using LRP.}
    \label{fig:PipeResults}
\end{figure*}

For a visual comparison between U-Net and our configuration with \textit{VGG-A one FC} and BMM, we refer to \autoref{fig:PipeResults}.
As can be seen in \autoref{fig:segmentation_unet}, U-Net generates many true positive crack segmentations, but gets distracted with strong brightness differences, for instance in the uppermost and lowest image.
Deposits depicted in the image are sometimes also mistakenly segmented as cracks, as can be seen in the second image from the bottom.
The LRP configuration is more robust against these issues, but the segmentations tend to be wider than the actual cracks, especially for narrow ones.

\begin{table*}[t]
\centering
\caption{Overview of the metrics for the sewer pipe cracks segmentation with our proposed LRP rule set (left) and the one proposed by \cite{Montavon2019} (right). On each side, the default VGG-A architecture and the architecture with only one fully connected layer are compared. For easier comparison, the U-Net results are listed on both sides.}
\label{tab:MetricsSewerPipes}
\begin{tabularx}{\textwidth}{@{}ccccLcccc@{}}
\cmidrule[\heavyrulewidth](r){1-4} \cmidrule[\heavyrulewidth](l){6-9}
\multicolumn{4}{c}{\textbf{LRP Ours}}                                                                                      &  & \multicolumn{4}{c}{\textbf{LRP Montavon}}                                                                                  \\ \cmidrule[\lightrulewidth](r){1-4} \cmidrule[\lightrulewidth](l){6-9} 
\textbf{VGG-A 128N}                 & \multicolumn{1}{l}{IoU} & \multicolumn{1}{l}{Precision} & \multicolumn{1}{l}{Recall} &  & \textbf{VGG-A 128N}                 & \multicolumn{1}{l}{IoU} & \multicolumn{1}{l}{Precision} & \multicolumn{1}{l}{Recall} \\ \cmidrule(lr){2-4} \cmidrule(l){7-9} 
\multicolumn{1}{c|}{Simple}         & 0.282                   & 0.381                         & 0.671                      &  & \multicolumn{1}{c|}{Simple}         & 0.292                   & 0.418                         & 0.614                      \\
\multicolumn{1}{c|}{GMM}            & 0.272                   & 0.331                         & 0.718                      &  & \multicolumn{1}{c|}{GMM}            & 0.249                   & 0.316                         & 0.689                      \\
\multicolumn{1}{c|}{BMM}            & 0.314                   & 0.469                         & 0.578                      &  & \multicolumn{1}{c|}{BMM}            & 0.303                   & 0.525                         & 0.483                      \\ \cmidrule[\lightrulewidth](r){1-4} \cmidrule[\lightrulewidth](l){6-9} 
\textbf{VGG-A one FC}               & IoU                     & \multicolumn{1}{l}{Precision} & \multicolumn{1}{l}{Recall} &  & \textbf{VGG-A one FC}               & IoU                     & \multicolumn{1}{l}{Precision} & \multicolumn{1}{l}{Recall} \\ \cmidrule(lr){2-4} \cmidrule(l){7-9} 
\multicolumn{1}{c|}{Simple}         & 0.297                   & 0.381                         & 0.727                      &  & \multicolumn{1}{c|}{Simple}         & 0.340                   & 0.464                         & 0.669                      \\
\multicolumn{1}{c|}{GMM}            & 0.270                   & 0.169                         & 0.782                      &  & \multicolumn{1}{c|}{GMM}            & 0.281                   & 0.332                         & 0.785                      \\
\multicolumn{1}{c|}{BMM}            & 0.337                   & 0.318                         & 0.647                      &  & \multicolumn{1}{c|}{BMM}            & 0.321                   & 0.686                         & 0.488                      \\ \cmidrule[\heavyrulewidth](r){1-4} \cmidrule[\heavyrulewidth](l){6-9} 
\multicolumn{1}{c|}{\textbf{U-Net}} & 0.321                   & 0.450                         & 0.797                      &  & \multicolumn{1}{c|}{\textbf{U-Net}} & 0.321                   & 0.565                         & 0.797                      \\ \cmidrule[\heavyrulewidth](r){1-4} \cmidrule[\heavyrulewidth](l){6-9} 
\end{tabularx}
\end{table*}

\vspace{\verticalSpacePreSections}
\section{\uppercase{Conclusion}}
\label{sec:conclusion}
\vspace{\verticalSpacePostSections}
We presented a method to circumvent the requirement of a pixel-wise labeling in order to train a neural network to accomplish this demanding task.
In order to demonstrate the applicability of our approach, we compare different configurations against the established U-Net architecture and achieve comparable results using two different datasets.
Thereby, we show that the output of the Layer-wise Relevance Propagation (LRP) can be exploited to generate pixel-wise segmentation masks.

\begin{figure}[h]
     \centering
     \begin{subfigure}[b]{\linewidth}
        \centering
        \includegraphics[width=0.32\textwidth]{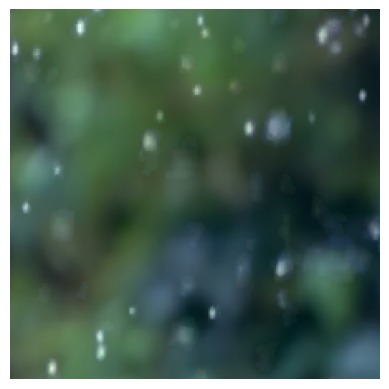}
        \includegraphics[width=0.32\textwidth]{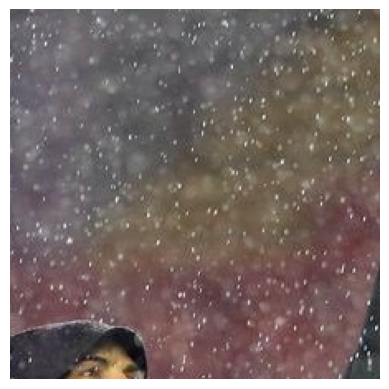}
        \includegraphics[width=0.32\textwidth]{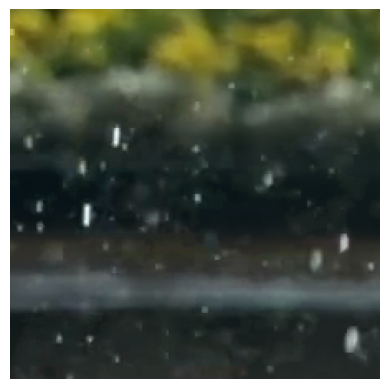}
    \end{subfigure}
    \begin{subfigure}[b]{\linewidth}
        \centering
        \includegraphics[width=0.32\textwidth]{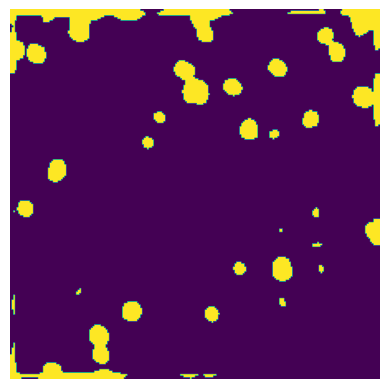}
        \includegraphics[width=0.32\textwidth]{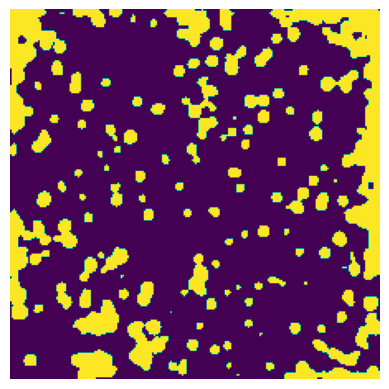}
        \includegraphics[width=0.32\textwidth]{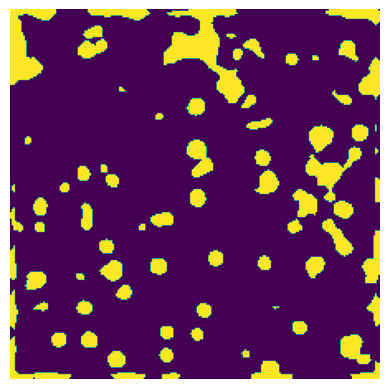}
    \end{subfigure}
    \caption{Some early results of rain streak segmentation on natural images.\label{fig:samples_rain}}        
\end{figure}

An interesting further research direction could be the extension to multi-label segmentation.
Currently, we also try to apply the proposed solution to the challenging problem of rain streak segmentation in natural images, as it is a very demanding task to generate a sufficient amount of training data and therefore is an ideal application area for our proposed method. Some early results of our work can be seen in \autoref{fig:samples_rain}.

\vfill

\section*{\uppercase{Acknowledgements}}
\label{sec:acknow}
\vspace{\verticalSpacePostSections}
This work has partly been funded by the German Federal Ministry of Economic Affairs and Energy under grant number 01MT20001D (Gemimeg), the Berlin state ProFIT program under grant number 10174498 (BerDiBa), and the German Federal Ministry of Education and Research under grant number 13N13891 (AuZuKa).

\bibliographystyle{apalike}
{\small
\bibliography{bibliography}}

\newpage


\end{document}